\documentclass[11pt,a4paper]{article}
\usepackage[hyperref]{emnlp2018}
\usepackage{times}
\usepackage{tabularx}
\usepackage{booktabs}
\usepackage{latexsym}
\usepackage{tcolorbox}
\usepackage{graphicx}
\usepackage{subcaption}
\usepackage{graphics}
\usepackage{adjustbox}
\usepackage{amsmath}
\usepackage{multirow}
\usepackage{multicol}
\usepackage{pbox}
\usepackage{xfrac}

\usepackage[colorinlistoftodos,textsize=scriptsize]
{todonotes}
\usepackage{multirow}
\usepackage{stfloats} 

\usepackage{url}

\aclfinalcopy 
\def\aclpaperid{1576} 

\newtcbox{\catbox}[1][white]{on line,
arc=3pt,colback=#1!50!black,colframe=#1!50!black,boxrule=1pt,
boxsep=0pt,left=1pt,right=1pt,top=1pt,bottom=1pt}

\newcommand{\stag}[1]{$^{\text{\color{gray}#1}}$}
\newcommand{\tworow}[1]{\multirow{2}{*}{#1}}

\title{What can we learn from Semantic Tagging?}

\author{Mostafa Abdou$^{\dagger}$$^{\star}$, Artur Kulmizev$^{\star}$, Vinit Ravishankar$^\S$, Lasha Abzianidze$^{\star}$, \and Johan Bos$^{\star}$\\ abdou@di.ku.dk, a.kulmizev@student.rug.nl, vinit.ravishankar@gmail.com, \{l.abzianidze, johan.bos\}@rug.nl \\ CLCG, University of Groningen$^{\star}$ \\ CoAStaL DIKU, University of Copenhagen$^{\dagger}$\\
LTG, University of Oslo$^{\S}$\\}

\date{}

\begin{document}
\maketitle
%
\begin{abstract}
We investigate the effects of multi-task learning using the recently introduced task of semantic tagging. We employ semantic tagging as an auxiliary task for three different NLP tasks: part-of-speech tagging, Universal Dependency parsing, and Natural Language Inference. We compare full neural network sharing, partial neural network sharing, and what we term the \textit{learning what to share} setting where negative transfer between tasks is less likely. Our findings show considerable improvements for all tasks, particularly in the \textit{learning what to share} setting, which shows consistent gains across all tasks. 
\end{abstract}

\section{Introduction}
\label{sec:intro}

Multi-task learning (MTL) is a recently resurgent approach to machine learning in which multiple tasks are simultaneously learned. By optimising the multiple loss functions of related tasks at once, multi-task learning models can achieve superior results compared to models trained on a single task. The key principle is summarized by \citet{caruana1998multitask} as ``MTL improves generalization by leveraging the domain-specific information contained in the training signals of related tasks". Neural MTL has become an increasingly successful approach by exploiting similarities between Natural Language Processing (NLP) tasks \citep{collobert2008unified,sogaard2016deep,plank2016multilingual}. Our work builds upon \citet{bjerva2016semantic}, who demonstrate that employing semantic tagging as an auxiliary task for Universal Dependency \citep{mcdonald2013universal} part-of-speech tagging can lead to improved performance.

The objective of this paper is to investigate whether learning to predict lexical semantic categories can be beneficial to other NLP tasks. 
To achieve this we augment single-task models (ST)\footnote{We replicate models which perform at or close to the state-of-the-art. Our choice of models is based on replicability. } with an additional classifier to predict semantic tags and jointly optimize for both the original task and the auxiliary semantic tagging task. Our hypothesis is that learning to predict semantic tags as an auxiliary task can improve performance of single-task systems. We believe that this is, among other factors, due to the following:

\begin{itemize}
 \setlength{\itemsep}{0.5mm}
  \setlength{\parskip}{.1mm}
  \setlength{\parsep}{0mm}
\item Providing the main task's model with a useful
 inductive bias, encouraging it to prefer representations that lead to semantically plausible hypotheses over those that are not.

\item Putting the focus of the main task model's attention on features that actually matter by providing additional evidence for the relevance or irrelevance of those features.

\item Reducing the risk of overfitting by minimizing the model's \textit{Rademacher Complexity}\footnote{The ability to fit random noise.} Representations which are learned for multiple tasks have been shown to generalize better \citep{baxter2000model}. 
\end{itemize}

We test our hypothesis on three disparate NLP tasks: (i) Universal Dependency part-of-speech tagging (UPOS), 
(ii) Universal Dependency parsing (UD DEP), a complex syntactic task; and (iii) Natural Language Inference (NLI), a complex task requiring deep natural language understanding.

\begin{figure*}[ht]
    \centering
    \includegraphics[scale=0.36]{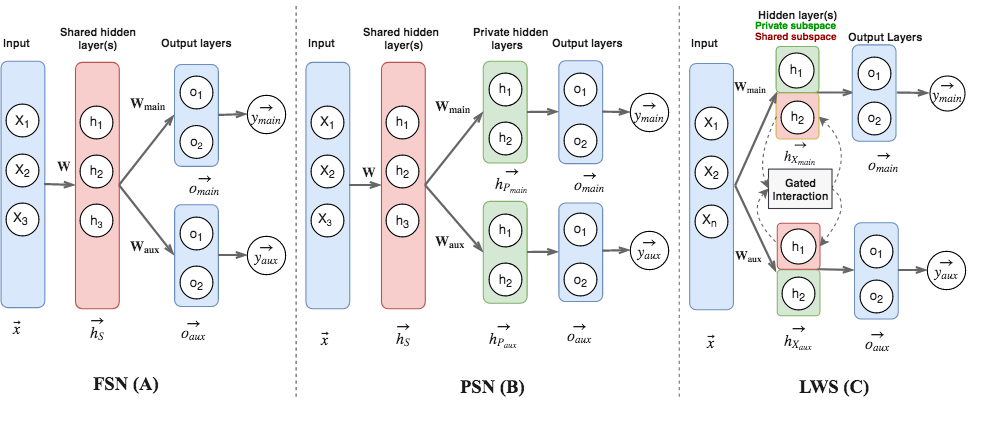}
    \vspace{-4mm}
    \caption{Our three multi-task learning settings: (A) fully shared networks, (B) partially shared networks, and (C) \textit{Learning What to Share}. Layers are mathematically denoted by vectors and the connections between them, represented by arrows, are mathematically denoted by matrices of weights. $S$ indicates a shared layer, $P$ a private layer, and $X$ a layer with shared and private subspaces.}
    \label{fig:mtlset}
\end{figure*}

\section{Background and Related work}
\subsection{Semantic Tagging}
Semantic tagging \cite{bjerva2016semantic,AbzianidzeBos2017IWCS} is the task of assigning language-neutral semantic categories to words. It is designed to overcome a lack of semantic information syntax-oriented part-of-speech tagsets, such as the Penn Treebank tagset\citep{marcus1993building}, usually have. 
Such tagsets exclude important semantic
distinctions, such as negation and modals, types of quantification, named entity types, and the contribution of verbs to tense, aspect, or event. 

The semantic tagset is language-neutral, abstracts over part-of-speech and named-entity classes, and includes fine-grained semantic information. The tagset consists of 80 semantic tags grouped in 13 coarse-grained classes.
The tagset originated in the Parallel Meaning Bank (PMB) project \citep{abzianidze2017parallel}, where it contributes to compositional semantics and cross-lingual projection of semantic representations.
Recent work has highlighted the utility of the tagset as a conduit for evaluating the semantics captured by vector representations \citep{belinkov2018evaluating}, or employed it in an auxiliary tagging task \citep{bjerva2016semantic}, as we do in this work.

\subsection{Learning What to Share}
Recently, there has been an increasing interest in the development of models which are trained to learn what to (and what not to) share  between a set of tasks, with the general aim of preventing negative transfer when the tasks are not closely related \citep{meyerson2017beyond, ruder2017sluice,lu2017fully,misra2016cross}. 
Our \textit{Learning What to Share} setting is based on this idea and closely related to \citet{liu2016recurrent}'s shared layer architecture. 

Specifically, a layer $\vec{h}_{X}$ which is shared between the main task and the auxiliary task is split into two subspaces: a shared subspace $\vec{h}_{X_{S}}$ and a private subspace $\vec{h}_{X_{P}}$. The interaction between the shared subspaces is modulated via a sigmoidal gating unit applied to a set of learned weights, as seen in Equations \eqref{eq:main} and \eqref{eq:aux} where $\vec{h}_{X_{S(main)}}$ and $\vec{h}_{X_{S(aux)}}$ are the main and auxiliary tasks' shared layers, $W_{a\rightarrow m}$ and $W_{m\rightarrow a}$ are learned weights, and $\sigma$ is a sigmoidal function.
\setlength{\abovedisplayskip}{5pt}
\setlength{\belowdisplayskip}{6pt}
\begin{align}
\vec{h}_{X_{S(main)}} &= \vec{h}_{X_{S(main)}} \sigma(\vec{h}_{X_{S(aux)}} W_{a\rightarrow m})\label{eq:main}\\
\vec{h}_{X_{S(aux)}} &= \vec{h}_{X_{S(aux)}} \sigma(\vec{h}_{X_{S(main)}}  W_{m\rightarrow a})\label{eq:aux}
\end{align}

Unlike \citet{liu2016recurrent}'s Shared-Layer Architecture, in our setup each task has its own shared subspace rather than one common shared layer. This enables the sharing of different parameters in each direction (i.e., from main to auxiliary task and from auxiliary to main task), allowing each task to choose what to learn from the other, rather than having ``one shared layer to capture the shared information for all the tasks'' as in \citet{liu2016recurrent}.  

\section{Multi-Task Learning Settings}
We implement three neural MTL settings, shown in Figure \ref{fig:mtlset}. They differ in the way the network's parameters are shared between the tasks:

\begin{itemize}
\setlength{\itemsep}{0.5mm}
\setlength{\parskip}{.1mm}
\setlength{\parsep}{0mm}
\item \textbf{Fully shared network (FSN):} All hidden layers are entirely shared among the tasks, each task has a separate output layer.
The transformation of our input vector $\vec{x}$ into a shared hidden layer $\vec{h}_S$ is described by Equation~\eqref{eq:fsn}: 
\setlength{\abovedisplayskip}{1pt}
\setlength{\belowdisplayskip}{4pt}
\begin{align}
\label{eq:fsn}
\vec{h}_S &= \sigma(\vec{x}W)
\end{align}
\item \textbf{Partially shared network (PSN):} A subset of hidden layers is shared among the tasks; each task has at least one private hidden layer and a separate output layer.
The transformation of a \textit{shared} hidden layer $\vec{h}_S$ into \textit{private} hidden layers, denoted by $\vec{h}_{P_{(main)}}$ and $\vec{h}_{P_{(aux)}}$ is described in Equations~\eqref{eq:psn1} and~\eqref{eq:psn2}.
\setlength{\abovedisplayskip}{1pt}
\setlength{\belowdisplayskip}{7pt}
\begin{align}
\label{eq:psn1}
\vec{h}_{P_{(main)}} &= \sigma(\vec{h_S}W_{(main)})\\
\label{eq:psn2}
\vec{h}_{P_{(aux)}} &= \sigma(\vec{h_S}W_{(aux)})
\end{align}

\item \textbf{Learning What to Share (LWS):} Each task has a dedicated set of hidden layers. For sharing, a hidden layer is split into a shared subspace and a private subspace. A gating unit modulates the transfer of information between the shared subspaces as shown in Equations~\eqref{eq:main} and~\eqref{eq:aux}. 
\end{itemize} 

\section{Data}

In the UPOS tagging experiments, we utilize the UD 2.0 English corpus \cite{nivre2017universal} for the POS tagging and the semantically tagged PMB release 0.1.0 (sem-PMB)%
\footnote{\url{http://pmb.let.rug.nl/data.php}}
for the MTL settings. Note that there is no overlap between the two datasets. Conversely, for the UD DEP and NLI experiments there is a complete overlap between the datasets of main and auxiliary tasks, i.e., each instance is labeled with both the main task's labels and semantic tags. We use the Stanford POS Tagger \cite{toutanova2003feature} trained on sem-PMB to tag the UD corpus and NLI datasets with semantic tags, and then use those assigned tags for the MTL settings of our dependency parsing and NLI models. 
We find that this approach leads to better results when the main task is only loosely related to the auxiliary task. The UD DEP experiments use the English UD 2.0 corpus, and the NLI experiments use the SNLI \citep{bowman2015large} and SICK-E\footnote{SICK-E refers to the entailment part of the SICK dataset.} datasets \citep{marelli2014sick}. The provided train, development, and test splits are used for all datasets. 
For sem-PMB, the silver and gold parts are used for training and testing respectively. 

\section{Experiments}

We run four experiments for each of the four tasks (UPOS, UD DEP, SNLI, SICK-E), one using the ST model and one for each of the three MTL settings. Each experiment is run five times, and the average of the five runs is reported. 
We briefly describe the ST models and refer the reader to the original work for further details due to a lack of space.%
\footnote{This applies to UD DEP and NLI only, as the POS tagger is not based on any one particular work.} 
For reproducibility, detailed diagrams of the MTL models for each task and their hyperparameters can be found in Appendix \ref{sec:supplementalA}.

\subsection{Universal Dependency POS Tagging}

Our tagging model uses a basic contextual one-layer bi-LSTM \cite{hochreiter1997long} that takes in word embeddings and produces a sequence of recurrent states which can be viewed as contextualized representations. The recurrent $r_n$ state from the bi-LSTM corresponding to each time-step $t_n$ is passed through a dense layer with a softmax activation to predict the token's tag. 

In each of the MTL settings a softmax classifier is added to predict a token's semantic tag and the model is then jointly trained on the concatenation of the sem-PMB and UPOS tagging data to minimize the sum of softmax cross-entropy losses of both the main (UPOS tagging) and auxiliary (semantic tagging) tasks.

\subsection{Universal Dependency Parsing}
We employ a parsing model that is based on \citeauthor{dozat2016deep} \citeyearpar{dozat2016deep}. The model's embeddings layer is a concatenation of randomly initialized word embeddings\footnote{This replaces the holistic word embeddings for frequent words in \citet{dozat2016deep}.} and character-based word representations added to pre-trained word embeddings, which are passed through a 4-layer stacked bi-LSTM. Unlike \citet{dozat2016deep}, our model jointly learns to perform UPOS tagging and parsing, instead of treating them as separate tasks. Therefore, instead of tag embeddings, we add a softmax classifier to predict UPOS tags after the first bi-LSTM layer. The outputs from that layer and the UPOS softmax prediction vectors are both concatenated to the original embedding layer and passed to the second bi-LSTM layer. The output of the last bi-LSTM is then used as input for four dense layers with a ReLU activation, producing four vector
representations: a word as a dependent
seeking its head; a word as a head seeking
all its dependents; a word as a dependent
deciding on its label; a
word as head deciding on the labels of its dependents. These representations are then passed to biaffine and affine softmax classifiers to produce a fully-connected labeled probabilistic dependency graph \citep{dozat2016deep}. 
Finally, a non-projective maximum spanning tree parsing algorithm \citep{chu1965shortest,edmonds1967optimum} is used to obtain a well-formed dependency tree.%
\footnote{This is recommended but not implemented by \citet{dozat2017stanford}.}

Similarly to UPOS tagging, an additional softmax classifier is used to predict a token's semantic tag in each of the MTL settings, as both tasks are jointly learned. In the FSN setting, the 4-layer stacked bi-LSTM is entirely shared. In the PSN setting the semantic tags are predicted from the second layer's hidden states, and the final two layers are devoted to the parsing task. In the LWS setting, the first two layers of the bi-LSTM are split into a private bi-LSTM$_{private}$ and a shared bi-LSTM$_{shared}$ for each of the tasks with the interaction between the shared subspaces being modulated via a gating unit. Then, two bi-LSTM layers that are devoted to parsing only are stacked on top.

\subsection{Natural Language Inference}
We base our NLI model on \citet{chen2017enhanced}'s Enhanced Sequential Inference Model which uses a bi-LSTM to encode the premise and hypothesis, computes a soft-alignment between premise and hypothesis' representations using an attention mechanism, and employs an inference composition bi-LSTM to compose
local inference information sequentially.%
\footnote{We do not implement the additional tree-LSTM model used in \citet{chen2017enhanced} as we focus on the effect of MTL with semantic tagging rather than on absolute performance.} 
The MTL settings are implemented by adding a softmax classifier to predict semantic tags at the level of the encoding bi-LSTM, with rest of the model unaltered. In the FSN setting, the hidden states of the encoding bi-LSTM are directly passed as input to the softmax classifier. In the PSN setting an earlier bi-LSTM layer is used to predict the semantic tags and the output from that is passed on to the encoding bi-LSTM which is stacked on top. This follows \citet{hashimoto2016joint}'s hierarchical approach. In the LWS setting, a bi-LSTM layer with private and shared subspaces is used for semantic tagging and for the ESIM model's encoding layer. In all MTL settings, the bi-LSTM used for semantic tagging is pre-trained on the sem-PMB data.

\section{Results and Discussion}

Results for all tasks are shown in Table \ref{tab:results}. In line with \citet{bjerva2016semantic}'s findings, the FSN setting leads to an improvement for UPOS tagging. POS tagging, a sequence labeling task, can be seen as the most closely related to semantic tagging, therefore negative transfer is minimal and the full sharing of parameters is beneficial. Surprisingly, the FSN setting also leads improvements for UD DEP. Indeed, for UD DEP, all of the MTL models outperform the ST model by increasing margins. For the NLI tasks, however, there is a clear degradation in performance. 

The PSN setting shows mixed results and does not show a clear advantage over FSN for UPOS and UD DEP. This suggests that adding task-specific layers after fully-shared ones does not always enable  sufficient task specialization. 
For the NLI tasks however, PSN is clearly preferable to FSN,
especially for the small-sized SICK-E dataset where the FSN model fails to adequately learn.  

\begin{table}[ht!]
\centering
\begin{adjustbox}{width=\linewidth}
\begin{tabular}{r|cc|c|c}
\toprule
\textbf{Model}
& \textbf{SNLI} & \textbf{SICK-E} & \textbf{UPOS} & \textbf{UD DEP}\\ 
\midrule
ST & 87.01  & 81.30 & 92.12 & 80.24 / 84.87\\ 
FSN &   84.96  & 56.69 & 92.95 &  81.03 / 85.54 \\
PSN &  87.08 & 77.92 & 92.34 &  80.92 / 85.81 \\
LWS& \textbf{87.51} & \textbf{84.57} & \textbf{95.54}& \textbf{81.39 / 86.00}\\
\bottomrule
\end{tabular}
\end{adjustbox}
\caption{Results for single-task models (ST), fully-shared networks (FSN), partially-shared networks (PSN), and \textit{learning what to share} (LWS).
All scores are reported as accuracy, except UD DEP for which we report LAS/UAS $F_1$ score. }
\label{tab:results}
\end{table}

As a sentence-level task, NLI is functionally dissimilar to semantic tagging. However, it is a task which requires deep understanding of natural language semantics and can therefore conceivably benefit from the signal provided by semantic tagging. Our results demonstrate that it is possible to 
leverage this signal given a selective sharing setup where negative transfer can be minimized. Indeed, for the NLI tasks, only the LWS setting leads to improvements over the ST models.%
\footnote{Demonstrative examples of the SNLI models' outputs can be found in Appendix \ref{sec:supplementalB}.} 
The improvement is larger for the SICK-E task which has a much smaller training set and therefore stands to learn more from the semantic tagging signal. For all tasks, it can be observed that the LWS models outperform the rest of the models. This is in line with our expectations with the findings from previous work \cite{ruder2017sluice,liu2016recurrent} that selective sharing outperforms full network and partial network sharing.

\section{Analysis}
\begin{figure*}[t!]
\centering
\begin{subfigure}[h!]{0.3\textwidth}
	\includegraphics[width=\textwidth]{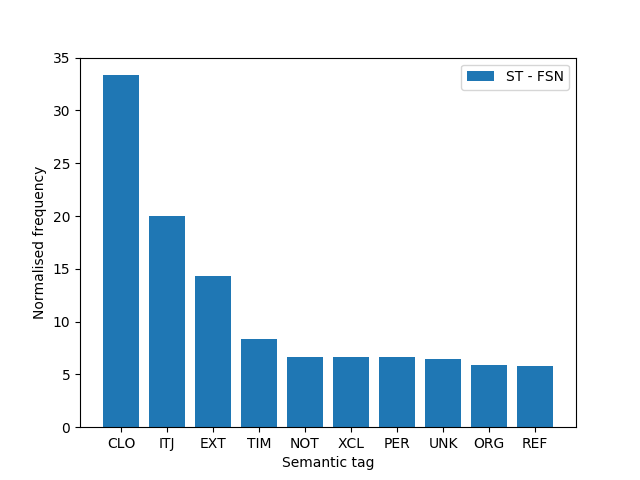}
    \caption{Single-task network}
\end{subfigure}
\begin{subfigure}[h!]{0.3\textwidth}
	\includegraphics[width=\textwidth]{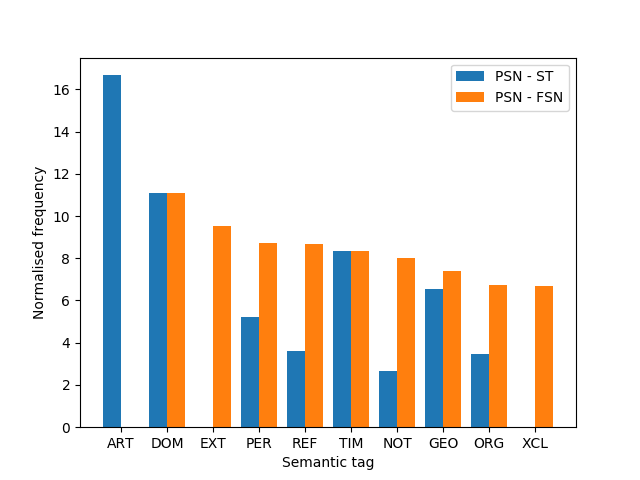}
    \caption{Partially shared network}
\end{subfigure}
\begin{subfigure}[h!]{0.3\textwidth}
	\includegraphics[width=\textwidth]{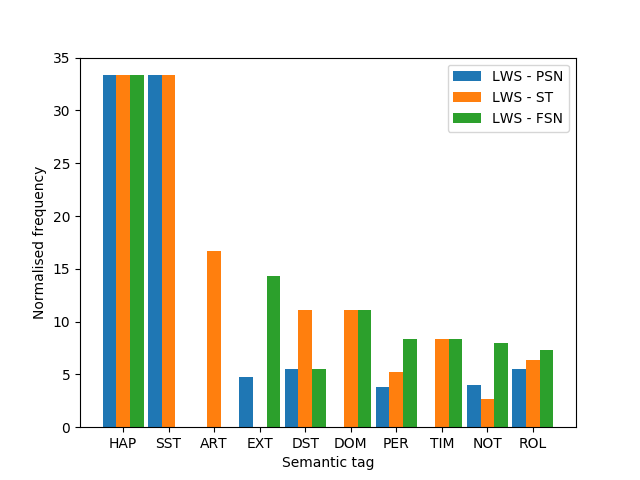}
    \caption{Learning what to share}
\end{subfigure}
\caption{Normalized semantic tag frequencies for all six sets of sentences. \texttt{X - Y} denotes the set of sentences correctly classified by model X but misclassified by model Y.}
\label{fig:nli_analysis}
\end{figure*}
In addition to evaluating performance directly, we attempt to qualify how semtags affect performance with respect to each of the SNLI MTL settings\footnote{We also provide a per-label report of the standard precision and recall metrics in Appendix \ref{sec:supplementalB}.}. 
\subsection{Qualitative analyses}
The fact that NLI is a sentence-level task, while semantic tags are word-level annotations presents a difficulty in measuring the effect of semantic tags on the systems' performance, as there is no one-to-one correspondence between a correct label and a particular semantic tag. We therefore employ the following method in order to assess the contribution of semantic tags. Given the performance ranking of all our systems --- $FSN < ST < PSN < LWS$ --- we make a pairwise comparison between the output of a superior system $S_{sup}$ and an inferior system $S_{inf}$. This involves taking the pairs of sentences that every $S_{sup}$ classifies correctly, but some $S_{inf}$ does not. Given that FSN is the worst performing system and, as such, has no `worse' system for comparison, we are left with six sets of sentences:  \textsc{ST-FSN}, \textsc{PSN-FSN}, \textsc{PSN-ST}, \textsc{LWS-PSN}, \textsc{LWS-ST}, and \textsc{LWS-FSN}. To gain insight as to where a given system $S_{sup}$ performs better than a given $S_{inf}$, we then sort each comparison sentence set by the frequency of semtags predicted therein, which are normalized by dividing by their frequency in the full SNLI test set.

We notice interesting patterns, visible in Figure~\ref{fig:nli_analysis}. Specifically, PSN appears markedly better at sentences with named entities (\texttt{ART}, \texttt{PER}, \texttt{GEO}, \texttt{ORG}) and temporal entities (\texttt{DOM}) than both ST and the FSN. Marginal improvements are also observed for sentences with negation and reflexive pronouns. The LWS setting continues this pattern, with additional improvements observable for sentences with the \texttt{HAP} tag for names of events, \texttt{SST} for subsective attributes, and the \texttt{ROL} tag for role nouns. 

\subsection{Contribution of semantic tagging}
To assess the contribution of the semantic tagging auxiliary task independent of model architecture and complexity we run three additional SNLI experiments --- one for each MTL setting --- where the model architectures are unchanged but the auxiliary tasks are assigned no weight (i.e. do not affect the learning). The results confirm our previous findings that, for NLI, the semantic tagging auxiliary task only improves performance in a selective sharing setting, and hurts it otherwise: i) the FSN system which had performed below ST improves to equal it and ii) the PSN and LWS settings both see a drop to ST-level performance. 
\section{Conclusions}

We present a comprehensive evaluation of MTL using a recently proposed task of semantic tagging as an auxiliary task. Our experiments span three types of NLP tasks and three MTL settings. The results of the experiments show that employing semantic tagging as an auxiliary task leads to improvements in performance for UPOS tagging and UD DEP in all MTL settings. For the SNLI tasks, requiring understanding of phrasal semantics, the selective sharing setup we term \textit{Learning What to Share} holds a clear advantage. Our work offers a generalizable framework for the evaluation of the utility of an auxiliary task.



\bibliography{emnlp2018}
\bibliographystyle{acl_natbib_nourl}

\appendix

\newpage
\clearpage

\section{MTL setting Diagrams, Preprocessing, and Hyperparameters}
\label{sec:supplementalA}
\subsection*{UPOS Tagging}
\begin{figure*}[bp]
\begin{subfigure}{\linewidth}
\includegraphics[width=\linewidth]{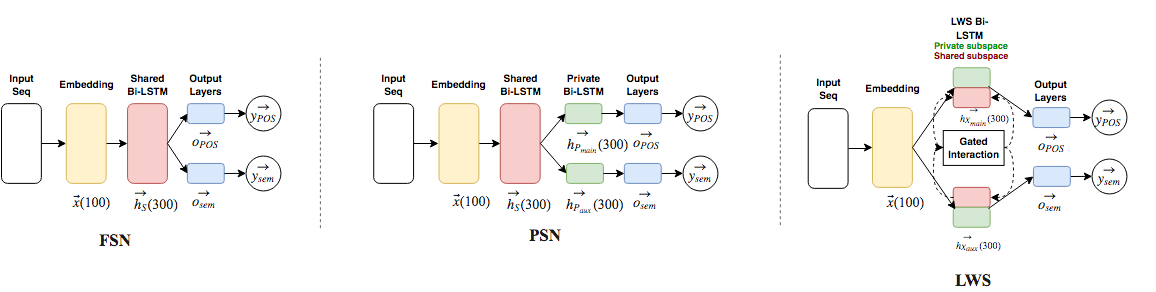}
\caption{UPOS}
\label{fig:upos}
\end{subfigure}
\vskip 0.1cm
\begin{subfigure}{0.5\linewidth}
\includegraphics[width=\linewidth]{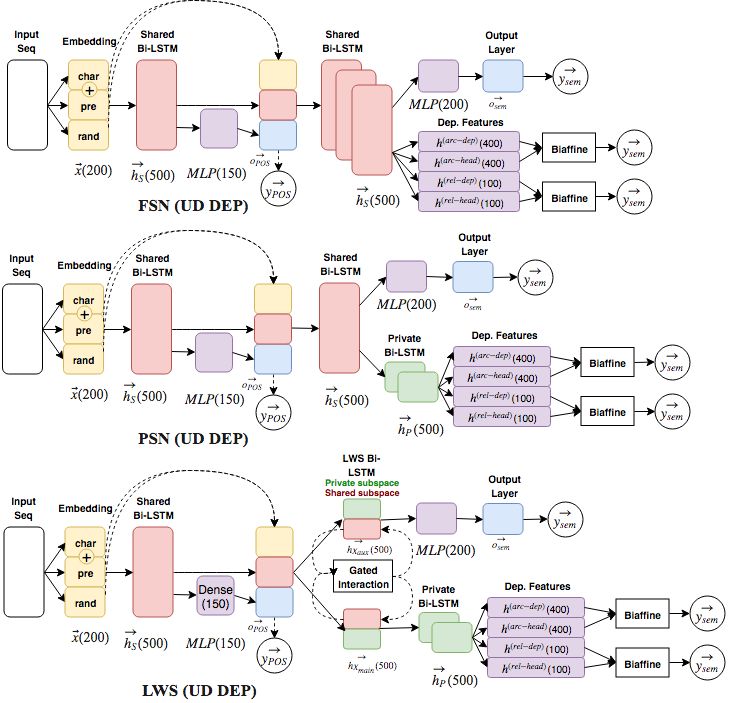}
\caption{UD DEP}
\label{fig:dep}
\end{subfigure}
\vrule
\hskip 0.2cm
\begin{subfigure}{0.5\linewidth}
\includegraphics[width=\linewidth]{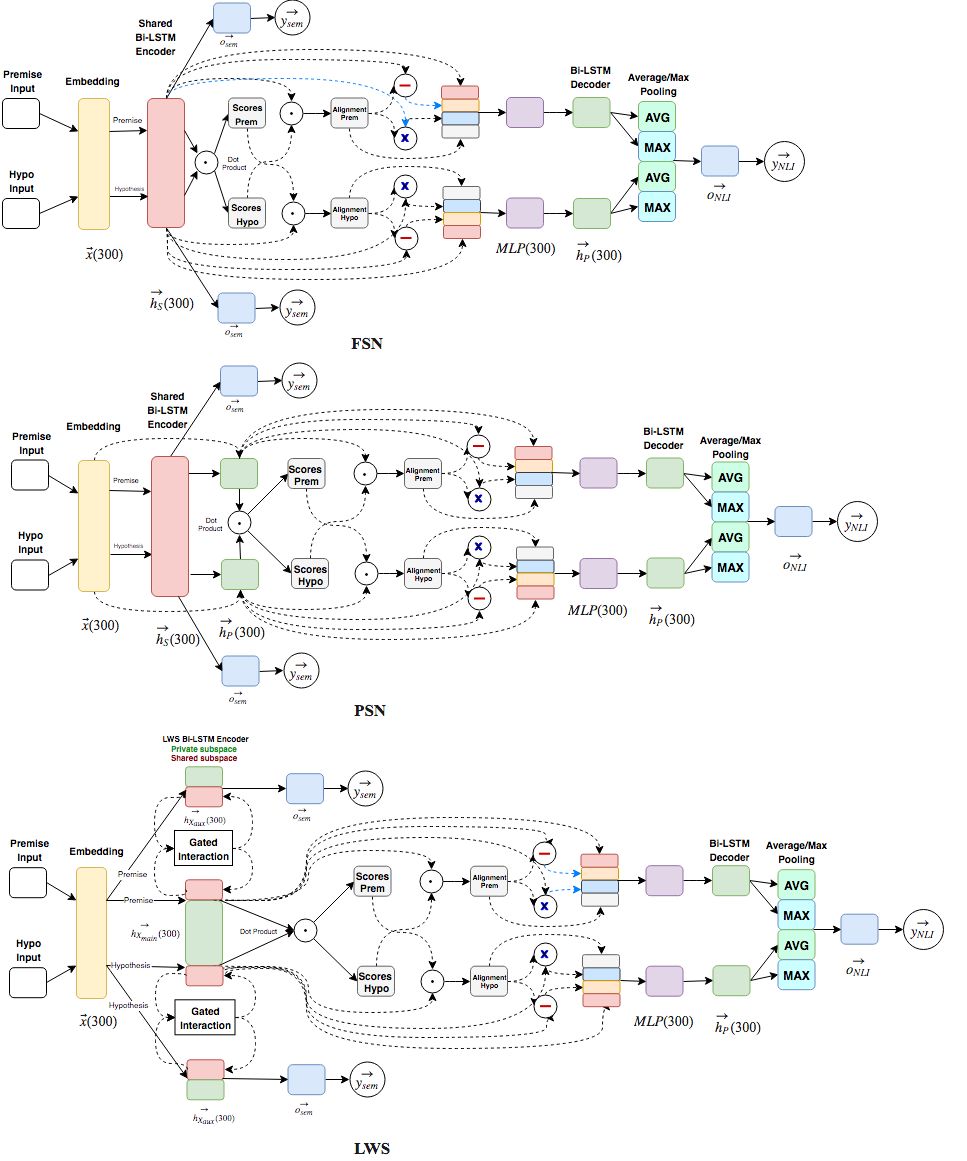}
\caption{NLI}
\label{fig:nli}
\end{subfigure}
\caption{The three MTL settings for each task. Layers dimensions are displayed in brackets.}
\end{figure*}

\autoref{fig:upos} shows the three MTL models used for UPOS. 
All hyperparameters were tuned with respect to loss on the English UD 2.0 UPOS validation set. We trained for 20 epochs with a batch size of 128 and optimized using Adam \citep{kingma2014adam} with a learning rate of $0.0001$. We weight the auxiliary semantic tagging loss with $\lambda$ = $0.1$. The pre-trained word embeddings we used are GloVe embeddings \citep{pennington2014glove} of dimension 100 trained on 6 billion tokens of Wikipedia 2014 and Gigaword 5. We applied dropout and recurrent dropout with a probability of $0.3$ to all bi-LSTMs.


\subsection*{UD DEP}
\autoref{fig:dep} shows the three MTL models for UD DEP. We use the gold tokenization. All hyperparameters were tuned with respect to loss on the English UD 2.0 UD validation set. We trained for 15 epochs with a batch size of 50 and optimized using Adam with a learning rate of  $2e-3$.  We weight the auxiliary semantic tagging loss with $\lambda$ = $0.5$. The pre-trained word embeddings we use are GloVe embeddings of dimension 100 trained on 6 billion tokens of Wikipedia 2014 and Gigaword 5. We applied dropout with a probability of $0.33$ to all bi-LSTM, embedding layers, and non-output dense layers.
\subsection*{NLI}
\autoref{fig:nli} shows the three MTL models for NLI. 
All hyperparameters were tuned with respect to loss on the SNLI and SICK-E validation datasets (separately). For the SNLI experiments, we trained for 37 epochs with a batch size of 128. For the SICK-E experiments, we trained for 20 epochs with a batch size of 8. Note that the ESIM model was designed for the SNLI dataset, therefore performance is non-optimal for SICK-E. For both sets of experiments: we optimized using Adam with a learning rate of $0.00005$; we weight the auxiliary semantic tagging loss with $\lambda$ = $0.1$; the pre-trained word embeddings we use are GloVe embeddings of dimension 300 trained on 840 billion tokens of Common Crawl; and we applied dropout and recurrent dropout with a probability of $0.3$ to all bi-LSTM, and non-output dense layers. 

\section{SNLI model output analysis}
\label{sec:supplementalB}

\autoref{tab:examples} shows demonstrative examples from the SNLI test set on which the \textit{Learning What to Share} (LWS) model outperforms the single-task (ST) model.
The examples cover all possible combinations of entailment classes. 
\autoref{tab:semtags} explains the relevant part of the semantic tagset. \autoref{tab:fscore} shows the per-label precision and recall scores.

\begin{table*}[bp!]
\centering
{\footnotesize
\begin{tabular}{@{}c@{~}lcc@{}}
\toprule
\multicolumn{2}{c}{\textbf{Premise-hypothesis pairs}} & \textbf{ST} & \textbf{LWS/GOLD}
\\
\bottomrule
\addlinespace[1mm]
\textbf{P}: & The\stag{DEF} gentleman\stag{CON} is\stag{NOW} speaking\stag{EXS} while\stag{SUB} the\stag{DEF} others\stag{ALT} are\stag{NOW} listening\stag{EXS}
& \tworow{N} & \tworow{E}
\\
\textbf{H}: & The\stag{DEF} man\stag{CON} is\stag{NOW} being\stag{EXS} given\stag{EXS} respect\stag{CON}
\\
\midrule
\textbf{P}: & Men\stag{CON} wearing\stag{EXG} hats\stag{CON} walk\stag{EXS} on\stag{REL} the\stag{DEF} street\stag{CON}
& \tworow{C} & \tworow{E}
\\
\textbf{H}: & The\stag{DEF} men\stag{CON} having\stag{EXS} hats\stag{CON} on\stag{REL} their\stag{HAS} head\stag{CON}
\\
\midrule
\textbf{P}: & Three\stag{QUC} men\stag{CON} in\stag{REL} orange\stag{IST} suits\stag{CON} are\stag{NOW} doing\stag{EXG} street\stag{CON} repairs\stag{CON} at\stag{REL} night\stag{CON}
& \tworow{N} & \tworow{C}
\\
\textbf{H}: & Three\stag{QUC} men\stag{CON} in\stag{REL} orange\stag{IST} suits\stag{CON} escaped\stag{EPS} from\stag{REL} prison\stag{CON}
\\
\midrule
\textbf{P}: & A\stag{DIS} toddler\stag{CON} sits\stag{ENS} on\stag{REL} a\stag{DIS} stone\stag{CON} wall\stag{CON} surrounded\stag{EXS} by\stag{REL} fallen\stag{EXS} leaves\stag{CON}
& \tworow{E} & \tworow{C}
\\
\textbf{H}: & An\stag{DIS} child\stag{CON} is\stag{NOW} throwing\stag{EXG} stones\stag{CON} at\stag{REL} a\stag{DIS} leaf\stag{CON} wall\stag{CON}
\\
\midrule
\textbf{P}: & An\stag{DIS} old\stag{IST} shoemaker\stag{CON} in\stag{REL} his\stag{HAS} factory\stag{CON}
& \tworow{C} & \tworow{N}
\\
\textbf{H}: & The\stag{DEF} shoemaker\stag{CON} is\stag{NOW} wealthy\stag{IST}
\\
\midrule
\textbf{P}: & A\stag{DIS} kid\stag{CON} slides\stag{CON} down\stag{IST} a\stag{DIS} yellow\stag{COL} slide\stag{CON} into\stag{REL} a\stag{DIS} swimming\stag{CON} pool\stag{CON}
& \tworow{E} & \tworow{N}
\\
\textbf{H}: & The\stag{DEF} kid\stag{CON} is\stag{NOW} playing\stag{EXS} at\stag{REL} the\stag{DEF} waterpark\stag{CON}
\\
\bottomrule
\end{tabular}
}
\caption{Examples of the entailment problems from SNLI which are incorrectly classified by the ST model but correctly classified by the LWS model.
Automatically assigned semantic tags are in superscript.}
\label{tab:examples}
\end{table*}

\begin{table}[h]
\centering

\scalebox{.67}{%
\begin{tabular}{@{}rl@{}}
\toprule
Tag category & Semantic tag with examples\\
\midrule
Anaphoric & \texttt{DEF}: definite; \emph{the, lo$^{IT}$, der$^{DE}$}\\
 & \texttt{HAS}: possessive pronoun; \emph{my, her}\\
\midrule
Attribute & \texttt{COL}: colour; \emph{red, crimson, light\_blue, chestnut\_brown}\\
& \texttt{QUC}: concrete quantity; \emph{two, six\textvisiblespace million, twice}\\
& \texttt{IST}: intersective; \emph{open, vegetarian, quickly}\\
& \texttt{REL}: relation; \emph{in, on, 's, of, after}\\
\midrule
Unnamed entity & \texttt{CON}: concept; \emph{dog, person}\\
\midrule
Logical & \texttt{ALT}: alternative \& repetitions; \emph{another, different, again}\\
& \texttt{DIS}: disjunction \& exist. quantif.; \emph{a, some, any, or}\\
\midrule
Discourse & \texttt{SUB}: subordinate relations; \emph{that, while, because}\\
\midrule
Events & \texttt{ENS}: present simple; \emph{{\scriptsize we} walk, {\scriptsize he} walks}
\\
& \texttt{EPS}: past simple; \emph{ate, went}\\
& \texttt{EXG}: untensed progessive; \emph{{\scriptsize is} running}\\
& \texttt{EXS}: untensed simple; \emph{{\scriptsize to} walk, {\scriptsize is} eaten, destruction}\\
\midrule
Tense \& aspect & \texttt{NOW}: present tense; \emph{is {\scriptsize skiing}, do {\scriptsize ski}, has {\scriptsize skied}, now}\\
\midrule
\end{tabular}
}
\caption{The list of semantic tags found in Table\,\ref{tab:examples}.}
\label{tab:semtags}
\end{table}

\begin{table}
\centering
\begin{tabular}{r|ccc}
\toprule
\multirow{2}{*}{\textbf{Model}} & \multicolumn{3}{|c}{\textbf{Label}}\\
& Entailment & Contradiction & Neutral\\
\midrule
\textbf{FSN} & 80.64/93.23 & 91.64/83.63 & 83.97/77.63\\
\textbf{ST} & 84.86/91.54 & 90.10/88.04 & 84.74/79.71\\
\textbf{PSN} & 84.08/92.70 & 91.17/88.63 & 85.96/79.15\\
\textbf{LWS} & 84.45/92.87 & 91.74/88.91 & 85.95/79.65\\
\bottomrule
\end{tabular}
\caption{Per-label precision (left) and recall (right) for all models.}
\label{tab:fscore}
\end{table}

\end{document}